\documentclass[12pt]{article}
\usepackage{amsmath,amssymb,graphicx,times,varioref}  
\usepackage{xcolor}
\usepackage{colortbl}
\title{Estimating the Operating Characteristics of Ensemble Methods}
\author{Anthony Gamst \and Jay-Calvin Reyes \and Alden Walker}
\date{March 2017}
\definecolor{dkgreen}{rgb}{0.0,0.5,0.0}
\newcommand{\red}[1]    {\textcolor{red}{#1}}

\newcommand{\green}[1]  {\textcolor{dkgreen}{#1}}

\newcommand{\gray}[1]   {\textcolor{gray}{#1}}
\begin{document} \maketitle
\begin{abstract}
  Suppose we would like to compare the performance of two binary
  classification algorithms.  If we have their receiver operating
  characteristic (ROC) curves, then one might conclude that the
  algorithm corresponding to the higher ROC curve is better.  This
  conclusion may not be correct due to variability in the ROC curves.
  Two sources of variability are randomness in the procedure used
  to train each classifier and randomness in the sample of feature 
  vectors that form the test data set.

  In this paper we present a technique for using the bootstrap to
  estimate the operating characteristics and their variability for
  certain types of ensemble methods.  Bootstrapping a model can
  require a huge amount of work if the training data set is large.
  Fortunately in many cases the technique lets us determine the effect
  of infinite resampling \emph{without actually refitting a single
    model}.  We apply the technique to the study of meta-parameter
  selection for random forests.  We demonstrate that alternatives to
  bootstrap aggregation and to considering $\sqrt{d}$ features to
  split each node, where $d$ is the number of features, can produce
  improvements in predictive accuracy.
\end{abstract}
\section{Introduction}

\begin{figure}
\begin{center}
  \includegraphics[height=3.0in]{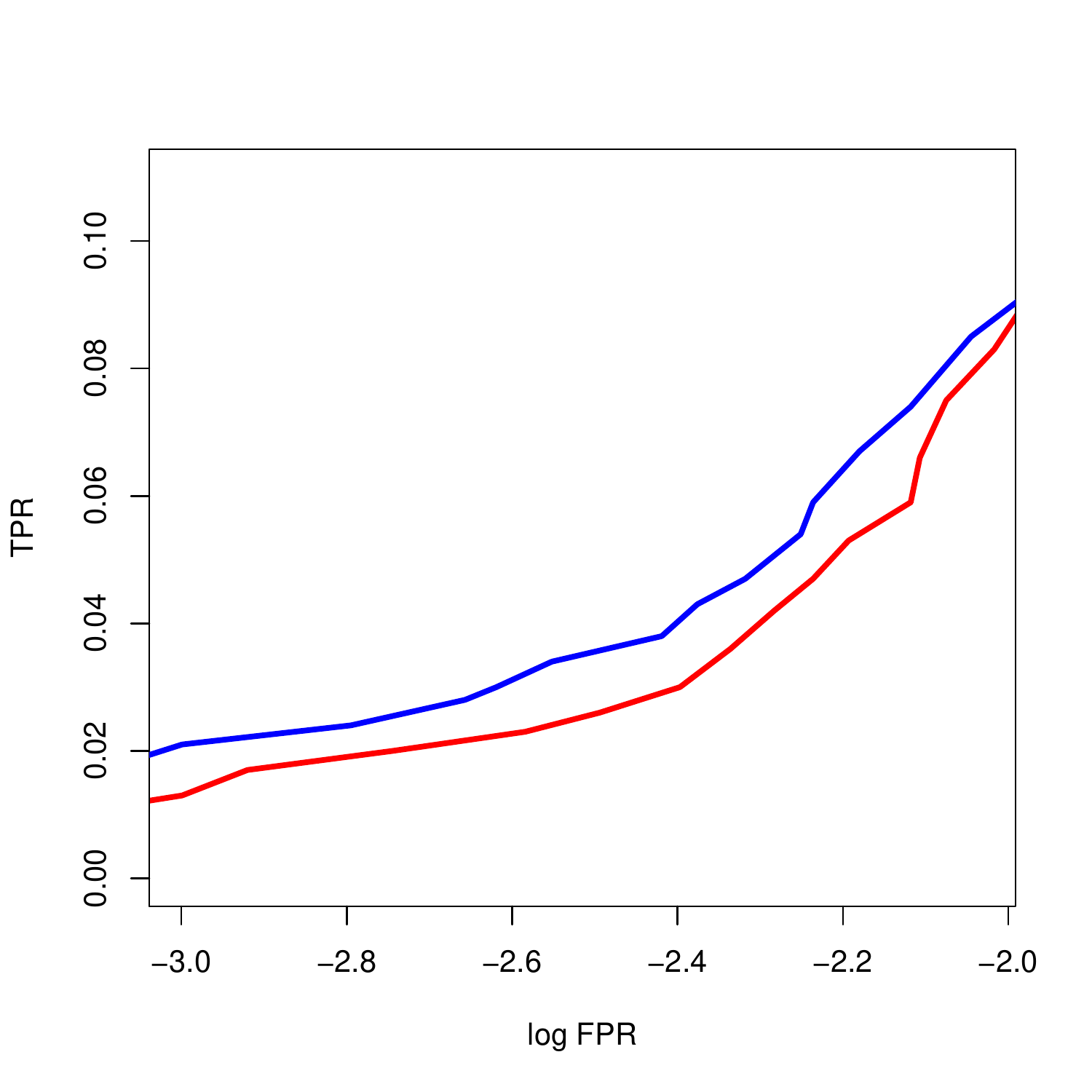}
  \caption{ROC curves for two binary classifiers.  Which algorithm is
    better?}
  \label{RF_variability}
\end{center}
\end{figure}

Suppose we would like to compare the performance of two binary
classification algorithms.  The receiver operating characteristic
(ROC) curve for each algorithm is shown in
Figure~\ref{RF_variability}.  A ROC curve is simply a plot of the true
positive rate (TPR) against the false positive rate (FPR) as the
discrimination threshold is varied.

The obvious conclusion is that since the blue ROC curve is higher, the
corresponding algorithm must be better.  In this case the conclusion
is not correct.  Both ROC curves were produced using the same
algorithm, the same training data set, and the same test data set.
The difference in the ROC curves is entirely due to randomness in
various choices made by the algorithm during training.  See
Section~\ref{variability_estimates} for a complete description of how
this example was generated.

An \emph{ensemble method} combines multiple weak learners to obtain
better predictive performance than could be obtained from any of the
individual learners alone \cite{wikipedia_ensemble_learning}.
For example, a \emph{random forest} uses an ensemble of decision
trees.  Each tree casts a vote and the forest combines the votes to
get a final prediction.

In this paper we describe a computationally efficient technique for
estimating the variability in ROC curves for ensemble methods that
base their decision on a vote among the weak classifiers.  The main
idea is to bootstrap the weak classifiers and to Poisson bootstrap
\cite{Chamandy} the feature vectors in the test set.  Note that the
technique has some ideas in common with previous unpublished work done
by Gamst and Goldschmidt.

Bootstrapping a model can require a huge amount of work if we need to
repeatedly fit the model to large data sets.  Fortunately in many
cases the technique described here lets us determine the effect of
doing infinite resampling \emph{without actually refitting a single
  model}.  

The rest of the paper is organized as follows.  In
Section~\ref{roc_curves} we describe the technique for generating a
ROC curve and for estimating its variability.  In
Section~\ref{examples} we describe various examples of applying the
new technique.  Finally in Section~\ref{conclusions} we give some
conclusions as well as directions for future work.

\section{Resampling the Weak Classifiers and the Test Set}
\label{roc_curves}

Let $X$ be the feature space and let $\mathcal{F}$ be a (possibly
infinite) set of binary classifiers\footnote{In this paper we assume
  that a (weak) binary classifier takes as input a feature vector and
  predicts either zero or one.  The theory developed in this section
  can be extended to weak classifiers that predict the probability of
  a feature vector belonging to class zero or one, but the formulas
  would become more complicated and the implementation would become
  more compute intensive.} on $X$.  Let $\mu_{\mathcal{F}}$ be a
measure on $\mathcal{F}$.  We construct an ensemble method $\gamma$ by
sampling $m$ weak classifiers $f_1, \ldots, f_m$ independently from
$\mathcal{F}$ according to $\mu_{\mathcal{F}}$.  Then for each $t \in
[0,m]$ the class label assigned to $x$ with threshold $t$ is:
\begin{align*}
  \hat{y}(x|\gamma,t) &= 1, \mbox{if} \sum_{i=1}^m f_i(x) \ge t, \mbox{and} \\
                      &= 0, \mbox{otherwise}.
\end{align*}
In other words, $\gamma$ predicts that $x$ is positive with threshold
$t$ if and only if at least $t$ of its weak classifiers predict that
$x$ is positive\footnote{There are other ways to combine the
  predictions of weak classifiers.  For example, we could use logistic
  regression or some other procedure to weight some of the weak classifiers
  more heavily than others. However, restricting to unweighted ensembles 
  has computational advantages that will be described later in this section.}.

Let $X \times Y$ be a (possibly infinite) space of labeled feature
vectors.
Let $\mu_{X \times Y}$ be a measure on $X \times Y$.  We evaluate the
ensemble method $F$ as follows.  First we sample $N$ labeled
feature vectors $(x_1,y_1),\ldots,(x_{N},y_{N})$ independently
from $X \times Y$ according to $\mu_{X \times Y}$.  Then for each $t
\in [0,m]$ we compute:
\begin{align*}
\mbox{FPR}(\gamma,t) &= \frac{|\{(x_j,y_j) : \hat{y}(x_j|\gamma,t) = 1 \mbox{\ and\ } y_j = 0\}|}{|\{(x_j,y_j) : y_j = 0\}|}. \\
\mbox{TPR}(\gamma,t) &= \frac{|\{(x_j,y_j) : \hat{y}(x_j|\gamma,t) = 1 \mbox{\ and\ } y_j = 1\}|}{|\{(x_j,y_j) : y_j = 1\}|}.
\end{align*}
For each $t$ we plot FPR($\gamma,t$) against TPR($\gamma,t$) to obtain
a ROC curve.

Note that a standard random forest fits this description.  Each
time a decision tree is produced, some randomness is introduced by
varying the bootstrapped sample of training feature vectors and the
features considered for splitting each node.  $\mathcal{F}$ is the
space of all possible decision trees that could have been produced
from the training data, and $\mu_{\mathcal{F}}$ gives the probability
of producing each decision tree.  We assume the $N$ test
feature vectors were drawn from a space of labeled feature vectors
according to some measure $\mu_{X \times Y}$.

The estimate of the ROC curve depends on which $f_i$ and which
$(x_j,y_j)$ we happened to sample.  Ideally we would like to know the
average ROC curve across all possible samples $\{f_i\}$ and
$\{(x_j,y_j)\}$.  Since compute and data limits prevent us from
obtaining arbitrarily many $f_i$ and $(x_j,y_j)$ we turn to resampling
techniques instead.

\subsection{Resampling the Weak Classifiers}

Let $p_{x_j} = \Pr[f(x_j) = 1]$ where $f$ is chosen randomly from
$\mathcal{F}$ according to $\mu_{\mathcal{F}}$.  Then
\begin{equation*} 
  \Pr[\hat{y}(x_j|\gamma,t) = 1] = \sum_{k=t}^m \binom{m}{k} p_{x_j}^k (1-p_{x_j})^{m-k}.
\end{equation*}
For each $x_j$ in the test set we can use the empirical estimate:
\begin{equation*}
  p_{x_j} \simeq \frac{1}{m} |\{f_i : f_i(x_j) = 1\}|.
\end{equation*}
The estimates of FPR and TPR then become:
\begin{align*}
\mbox{FPR}(\gamma,t) &= \frac{\sum_{j=1}^{N} \Pr[\hat{y}(x_j|\gamma,t)=1] \chi(y_j=0)}{|\{(x_j,y_j) : y_j = 0\}|}. \\
\mbox{TPR}(\gamma,t) &= \frac{\sum_{j=1}^{N} \Pr[\hat{y}(x_j|\gamma,t)=1] \chi(y_j=1)}{|\{(x_j,y_j) : y_j = 1\}|}.
\end{align*}
Here $\chi()$ is the indicator function that has value one when the
statement is true and zero otherwise.

The idea is to account for the variability from the choice of
$\{f_i\}$.  If a test feature vector $x_j$ received $k$ positive
votes, then it might have received fewer or more votes from a
different sample of $\{f_i\}$.  Thus $x_j$ has some probability of
satisfying a higher threshold than $k$ or of not satisfying a lower
threshold.  We can also estimate the variance in FPR($\gamma$) and
TPR($\gamma$) since they are produced by summing \emph{i.i.d.} random
variables.  The formula for the variance will be given in the next
section since it will also account for the variability from the choice
of the test set.

\subsection{Resampling the Test Set}

Given one test set of $N$ feature vectors we would like to
evaluate an ensemble method across many different test sets.  In the
absence of more data we can create a new test by sampling with
replacement from the original one.  To simplify calculations we will
perform a Poisson bootstrap on the test set\cite{Chamandy}. For each
$(x_j,y_j)$ in the original test set we will put $c_j$ copies in the
new test set where $c_j$ is a rate-1 Poisson random variable,
\emph{i.e.}, $\Pr[c_j = k] = e^{-1}/k!$.  Note that the new test set
has approximately $N$ feature vectors.  The quality of the
approximation improves as $N$ becomes large.  Now the estimates
of FPR and TPR are:
\begin{align*}
\mbox{FPR}(\gamma,t) &= \frac{\sum_{j=1}^{N} c_j \Pr[\hat{y}(x_j|\gamma,t)=1] \chi(y_j=0)}{|\{(x_j,y_j) : y_j = 0\}|}. \\
\mbox{TPR}(\gamma,t) &= \frac{\sum_{j=1}^{N} c_j \Pr[\hat{y}(x_j|\gamma,t)=1] \chi(y_j=1)}{|\{(x_j,y_j) : y_j = 1\}|}.
\end{align*}

Including $c_j$ in the above expressions doesn't change the mean,
but it does affect the variance.  Let $b_j$ be a random variable that
is 1 with probability $\Pr[\hat{y}(x_j|\gamma,t) = 1]$ and 0
otherwise.  Then
\begin{equation*}
  \mbox{Var}(c_j b_j) = [E(c_j)]^2 \mbox{Var}(b_j) + [E(b_j)]^2 \mbox{Var}(c_j) +
  \mbox{Var}(b_j)\mbox{Var}(c_j).
\end{equation*}
Let $q_j(t) = \Pr[\hat{y}(x_j|\gamma,t)) = 1]$.  Since $E(b_j)=q_j$,
$\mbox{Var}(b_j) = q_j(1-q_j)$, $E(c_j) = 1$, and $\mbox{Var}(c_j) =
1$ we get:
\begin{align*}
  \mbox{Var}(c_j b_j) &= q_j(1-q_j) + q_j^2 + q_j(1-q_j)\\
                      &= q_j + q_j(1-q_j).
\end{align*}
This represents the (unnormalized) contribution of the test feature vector
$x_j$ sampled with multiplicity $c_j$ to the variance of the FPR or
TPR.  Note that normalizing by ${|\{(x_j,y_j) : y_j = 0\}|}$ or
${|\{(x_j,y_j) : y_j = 1\}|}$ is no longer correct since the Poisson
resampling changes the number of test-set feature vectors in each class.
However, the error introduced by the normalization should be small if
$N$ is sufficiently large.  To estimate Var(FPR) and Var(TPR)
we assume the $b_j$ are independent so we can sum the variances over
$j$.

\subsection{Comparison of Techniques}

In unpublished work, Gamst and Goldschmidt resampled the test
vectors and computed the threshold $t$ corresponding to a particular
FPR.  They then averaged the thresholds produced from many samples.
This arrangement of the work forced them to either generate each
sample or to do complicated calculations with multinomial
coefficients.

The technique described in this paper resampled both the test
vectors and the weak classifiers.  It computed the FPR and TPR
corresponding to a particular threshold $t$.  The advantage of
arranging the work in this way was that the effect of taking
infinitely many samples from the training vectors and the weak
classifiers could be computed without actually generating any samples.
Furthermore, most of the computation did not depend on the number $m$
of weak classifiers in the ensemble.  Therefore the effect of varying
the size of the ensemble could be determined for a relatively small
amount of additional work.

Chamandy \emph{et. al.} \cite{Chamandy} described using the
Poisson bootstrap to determine the variability in a statistic computed
on an extremely large data set.  They did not focus on ensemble methods
and in particular they did not discuss resampling the weak classifiers
in an ensemble.

\section{Examples}
\label{examples}

In this section we give some examples of estimating the operating
characteristics of ensemble methods.

\subsection{Data}
\label{examples_data}

To demonstrate the technique, we wanted an interesting binary
classification problem related to real-world data with a relatively
large sample.  We also wanted the problem to be difficult but not
impossible.  With that in mind, we did the following.  First, we
obtained the original ImageNet~\cite{ImageNet} data set of about 1.2
million images each labeled with one of 1000 classes.  We took the
VGG16~\cite{VGG16} model provided by Keras~\cite{Keras} and removed
the top two layers.  Feeding an image into the resulting model
produces a feature vector of length 4096 which should provide
information about the image label.  Indeed, retraining the top two
layers of VGG16 (a dense ReLU layer of size 4096 followed by a dense
softmax layer of size 1000) quickly regains much of the accuracy of
the original VGG16 model.  To produce a binary classification problem,
we arbitrarily made labels 0--499 correspond to class 0 and labels
500-999 correspond to class 1.  Thus, our data consists of 1.2 million
4096-long feature vectors, each with a 0-1 label.



\subsection{Variability Estimates}
\label{variability_estimates}

\begin{figure}
\begin{center}
  \includegraphics[height=3.0in]{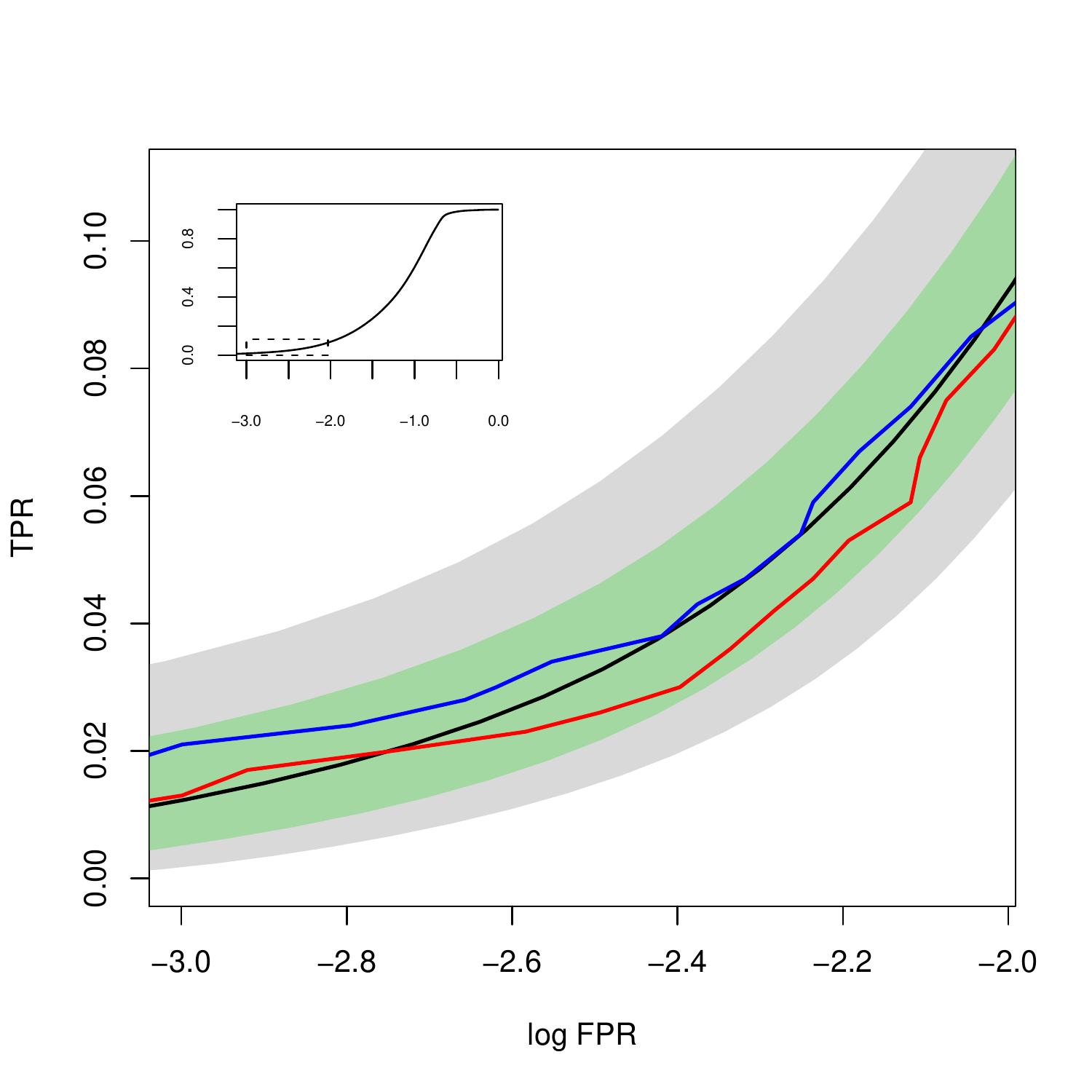}
  \caption{ROC curves for two random forest classifiers for the image
    data set.  Each random forest was trained on 10K images and tested
    on 10K images.  The shaded regions are approximate 95\% confidence
    bands \gray{with} and \green{without} Poisson bootstrapping the test
    data set.  The inset shows the full ROC curve. Throughout this paper
    the x-axis of the ROC curves is on a log-scale and the logs are
    taken base-10.}
  \label{RF_variability_estimate}
\end{center}
\end{figure}

In Figure~\ref{RF_variability} both ROC curves are for a random forest
with 256 trees where each tree has a maximum depth of 20.  Each forest
was trained on 10K images and then tested on 10K images.  The
difference between the red and blue ROC curves is due to each forest
making different random decisions when building its decision trees.
Note that for this example we intentionally used a small subset of the
image data to show the variability in the ROC curves.  In later
examples we will usually train the random forests on 1M images and
test them on 200K images.  In each example each tree has a maximum
depth of 20 unless otherwise indicated.

Figure~\ref{RF_variability_estimate} shows approximate 95\% confidence
bands for a random forest with 256 trees.  The gray confidence band
accounts for the variability from the choice of test data set, and the
green confidence band does not.  Both ROC curves from
Figure~\ref{RF_variability} lie within both confidence bands, so the
difference between the red and blue ROC curves is unsurprising.  For
the remaining examples in this paper we will \emph{not} account for
the variability from the choice of test data set since all of the
variations of random forest used the same training data set and test
data set.

The confidence bands in this example and in all subsequent examples
were constructed by the following procedure.  First, we built an
ensemble of 4096 trees.  Next we applied the technique described
earlier in this paper to compute the mean and variance for each point
on the ROC curve.  Finally, we applied the method described in section
4.3 of~\cite{hardle-anr} to produce the confidence bands, although
other approaches, including those described in Sun and Loader
\cite{sun_loader} could be used.

\subsection{Reduced Training Data}

\begin{figure}
\begin{center}
  \includegraphics[height=3.0in]{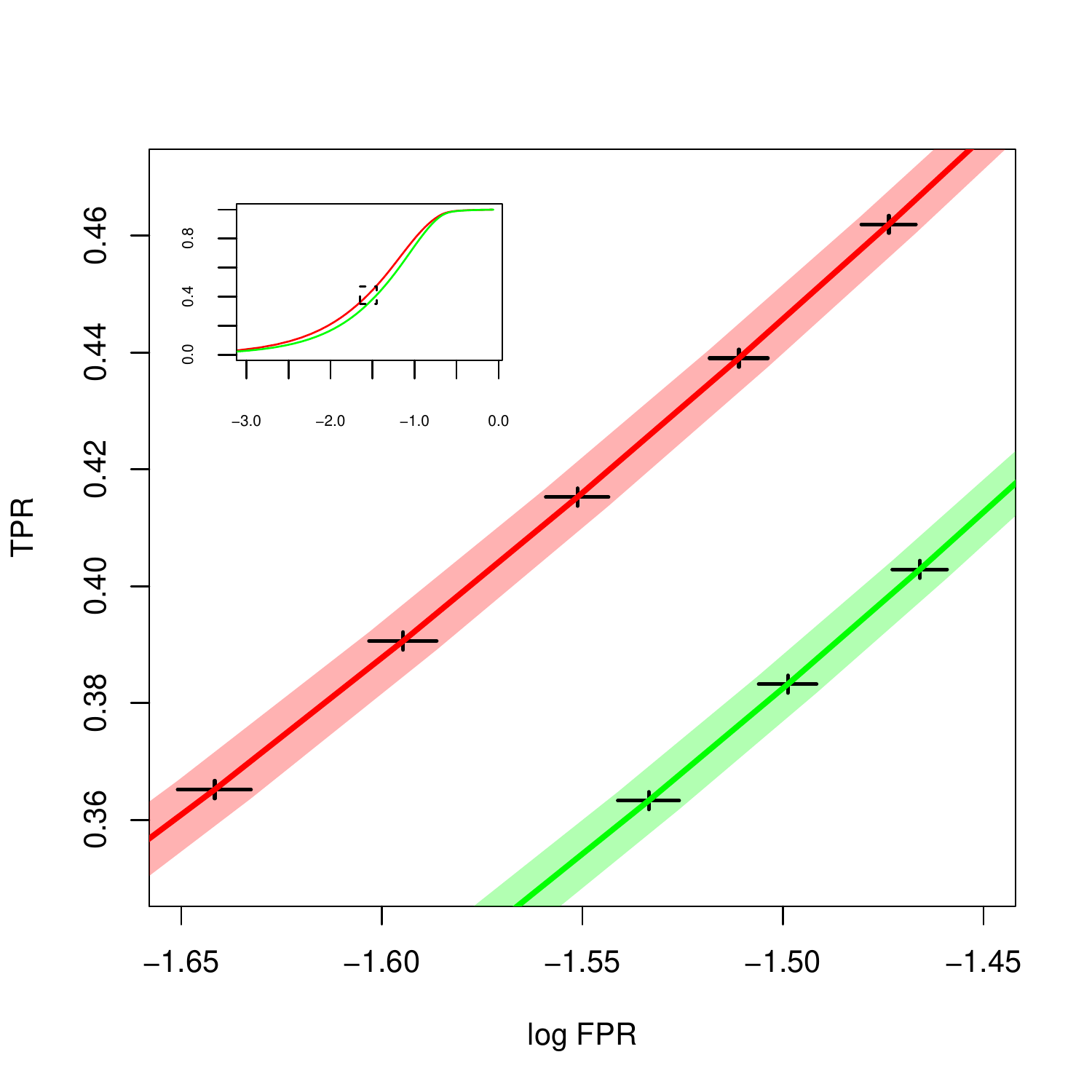}
  \caption{Confidence bands for random forests trained on \red{1M
      images} and \green{200K images}.  Also shown are two standard
    deviation confidence intervals (black) for log FPR and TPR at
    several points along the curve. The inset shows the full ROC
    curves. }
  \label{RF_less_training_data}
\end{center}
\end{figure}

It is well known that using more training data to build a classifier
usually improves its performance.  This is illustrated by
Figure~\ref{RF_less_training_data} which compares the operating
characteristic estimates for a random forest trained on 1M images and
a random forest trained on 200K images.  Reducing the amount of
training data causes the random forest to perform significantly worse.

\subsection{Fewer or More Trees}

\begin{figure}
\begin{center}
  \includegraphics[height=3.0in]{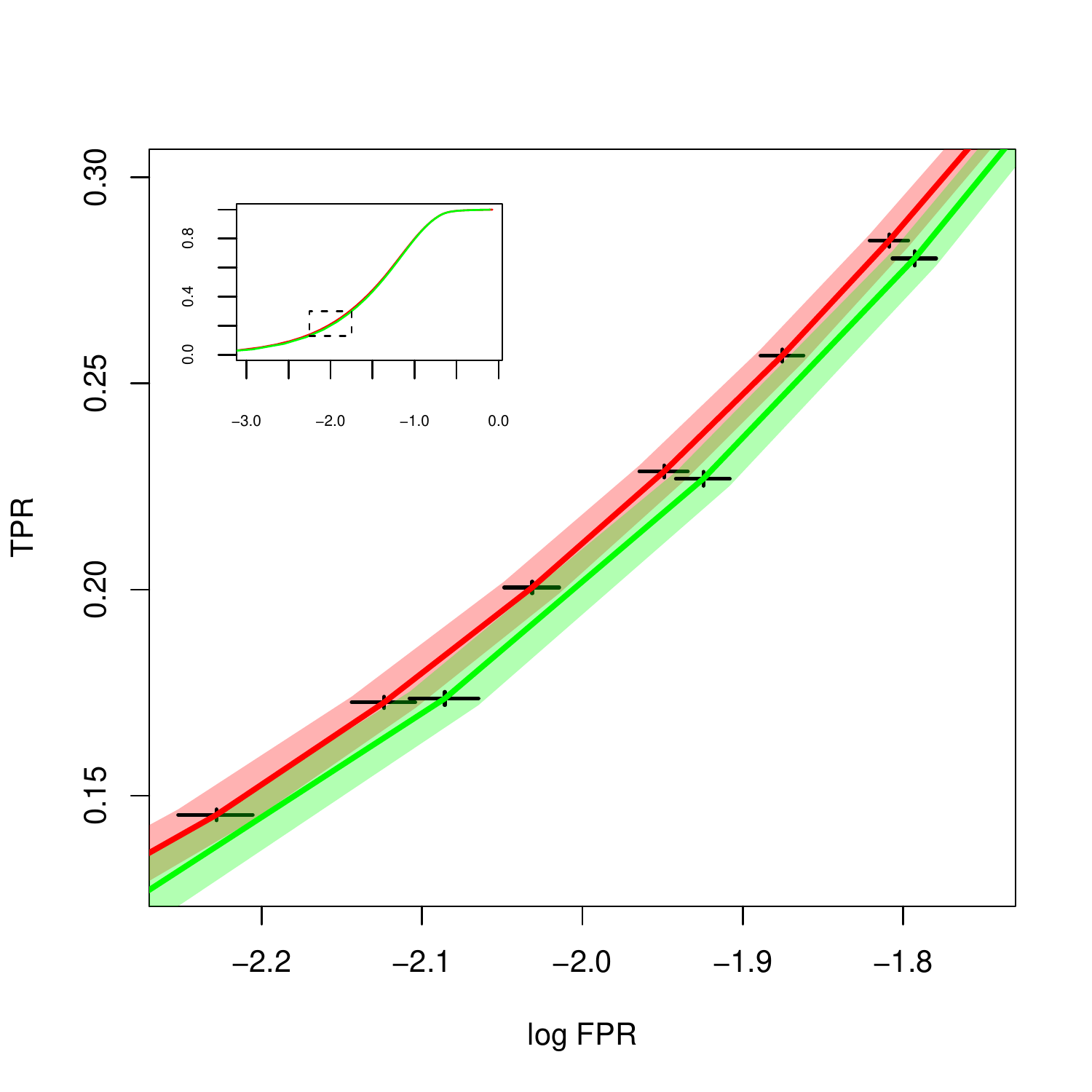}
  \caption{Confidence bands for random forests with \red{256 trees}
    and \green{128 trees}.  Also shown are two standard
    deviation confidence intervals (black) for log FPR and TPR at
    several points along the curve. The inset shows the full ROC
    curves.}
  \label{RF_fewer_trees}
\end{center}
\end{figure}

\begin{figure}
\begin{center}
  \includegraphics[height=3.0in]{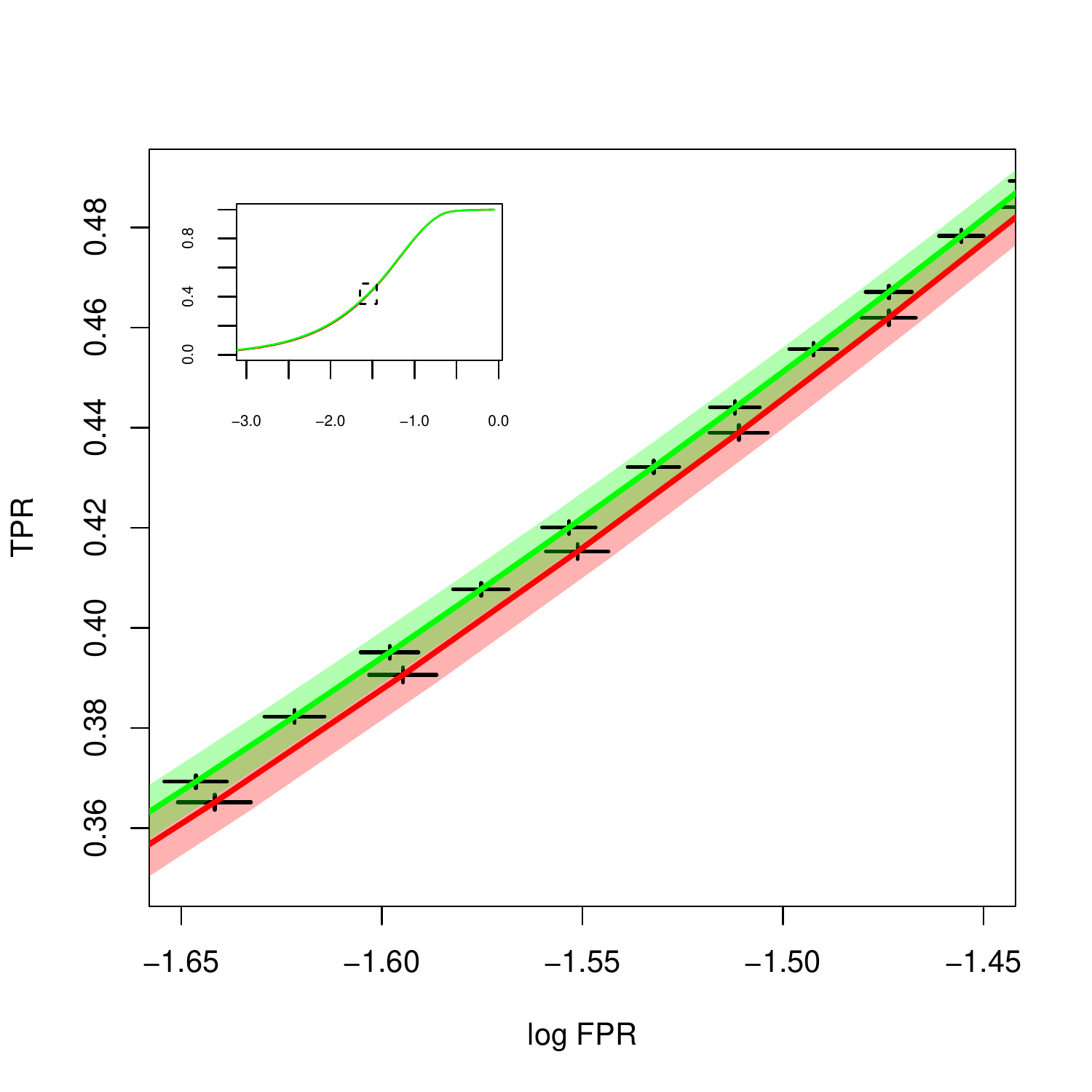}
  \caption{Confidence bands for random forests with \red{256 trees}
    and \green{512 trees}.  Also shown are two standard
    deviation confidence intervals (black) for log FPR and TPR at
    several points along the curve. The inset shows the full ROC
    curves. }
  \label{RF_more_trees}
\end{center}
\end{figure}

If a random forest does not have enough trees then we expect its
performance to be worse.  The theory of random forests asserts that
adding more trees cannot (asymptotically) hurt performance.
Figure~\ref{RF_fewer_trees} compares the confidence bands for 128-tree
and 256-tree forests.  Figure~\ref{RF_more_trees} compares the
confidence bands for 256-tree and 512-tree forests.  In both cases
when the number of trees is doubled the performance of the forest
improves.  The gain is smaller when going from 256 trees to 512 trees
suggesting that the returns from adding more trees are starting to
diminish.

\subsection{Bounded Tree Depth}

\begin{figure}
\begin{center}
  \includegraphics[height=3.0in]{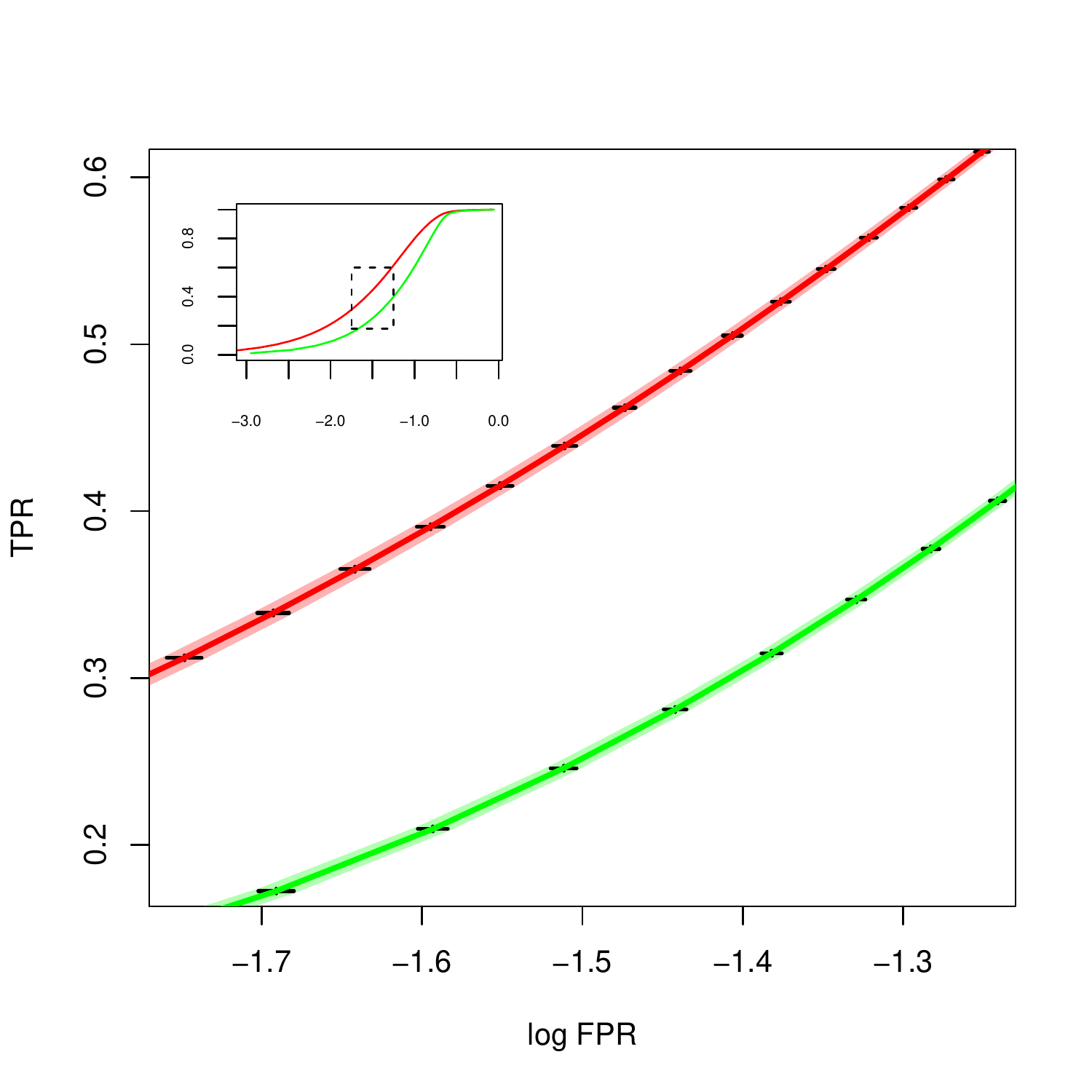}
  \caption{Confidence bands for random forests whose trees have
    \red{maximum depth 20} and \green{maximum depth 10}.  Also shown
    are two standard deviation confidence intervals (black) for log
    FPR and TPR at several points along the curve. The inset shows the
    full ROC curves. }
  \label{RF_bounded_tree_depth}
\end{center}
\end{figure}

The {\tt sklearn} implementation of a random forest gives the user the
option to bound the depths of the trees.  This reduces the size of
each tree, potentially reducing the training time, testing time, and
storage requirements for the forest.  It may also reduce overfitting
to the training data.

Figure~\ref{RF_bounded_tree_depth} compares the operating
characteristic estimates for random forests with maximum tree depths
of 10 and 20.  For this data set reducing the maximum tree depth hurts
the performance of the forest.

This is likely to be a general phenomenon. Restricting the depth of a
decision tree will increase its approximation error. An $s$-split tree
for example can only approximate functions of $s$ or fewer
features. Averaging has no effect on approximation error, although it
does reduce estimation error. The average of noisy estimates has
variance less than that of any one of the estimates; how much less
depends on how correlated the various estimates are (and the size of
the weights used to compute the average). So, restricted depth trees
should perform worse than their expanded cousins up to the point where
the approximation error is ``good enough''.  This trade-off is more
easily balanced by setting the minimum number of observations allowed
in any leaf node; that is, refusing to split if the suggested split
would produce fewer than $k_n$ observations in either daughter node (see
below). This is because splits defined over nodes with few
observations are likely to be spurious, whereas the fitted trees may
need to be both deep and narrow, splitting off small subsets at each
level.  Regardless, cross-validation and other techniques can be used
to select approximately optimal values for such meta-parameters.

\subsection{No Bootstrapping Training Data}

\begin{figure}
\begin{center}
  \includegraphics[height=3.0in]{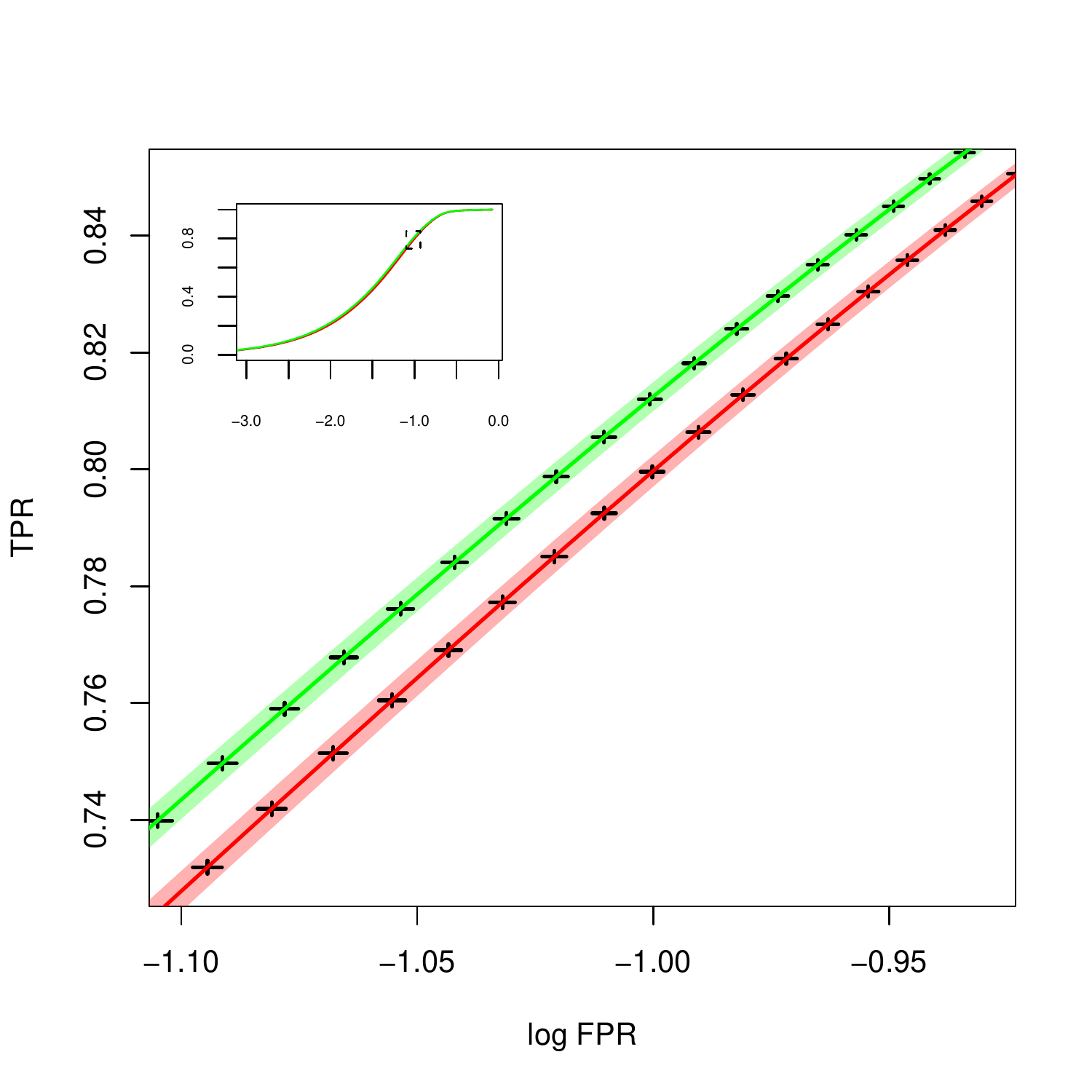}
  \caption{Confidence bands for random forests trained \red{with
      bootstrapping} and \green{without bootstrapping} of the training
    data set.  Also shown are two standard deviation confidence
    intervals (black) for log FPR and TPR at several points along the
    curve. The inset shows the full ROC curves. }
  \label{RF_no_bootstrap}
\end{center}
\end{figure}

A random forest uses two ideas to reduce the correlation between
trees.  First, a bootstrapped sample of the training feature vectors
are used to create each tree.  Second, a random subset of the features
is considered when splitting each node.

As a thought experiment consider a random forest with a single tree.
In this case it's not clear that we need to bootstrap the training
vectors when building the tree.  Furthermore, if we believe that
bootstrapping is harmful for a small forest and beneficial for a large
forest, then there should be some crossover point.  Can we observe
this crossover point, and is it smaller or larger than the forests
that we actually produce?

The original purpose of the bootstrapping and averaging was to
stabilize the predictions from individual trees. The structure of and
classifications produced by individual trees are very sensitive to the
specifics of the training data, in the sense that small changes to the
training data set can produce very different trees.  Bootstrap samples
can be seen as small perturbations to the training data and it makes
sense to trust the average tree over all perturbations than any single
tree. On the other hand, this effect should vanish as the size of the
training data set increases, provided the individual trees don't grow
too fast as the sample size increases. Regardless, one could consider
sub-sampling or different (high-entropy) weighting schemes as
alternatives to the bootstrap, using cross-validation or some other
procedure to choose between the alternatives. For example,
bootstrapping corresponds roughly with re-weighting the observations
according to samples from a very specific distribution. This
distribution can be embedded in a one-parameter exponential family and
cross-validation can be used to select the optimal value for this
hyper-parameter.

Figure~\ref{RF_no_bootstrap} compares the operating characteristic
estimates for random forests built with and without bootstrapping the
training set vectors.  The forest built without bootstrapping performs
better.

\subsection{Considering More Features to Split Each Node}

\begin{figure}
\begin{center}
  \includegraphics[height=3.0in]{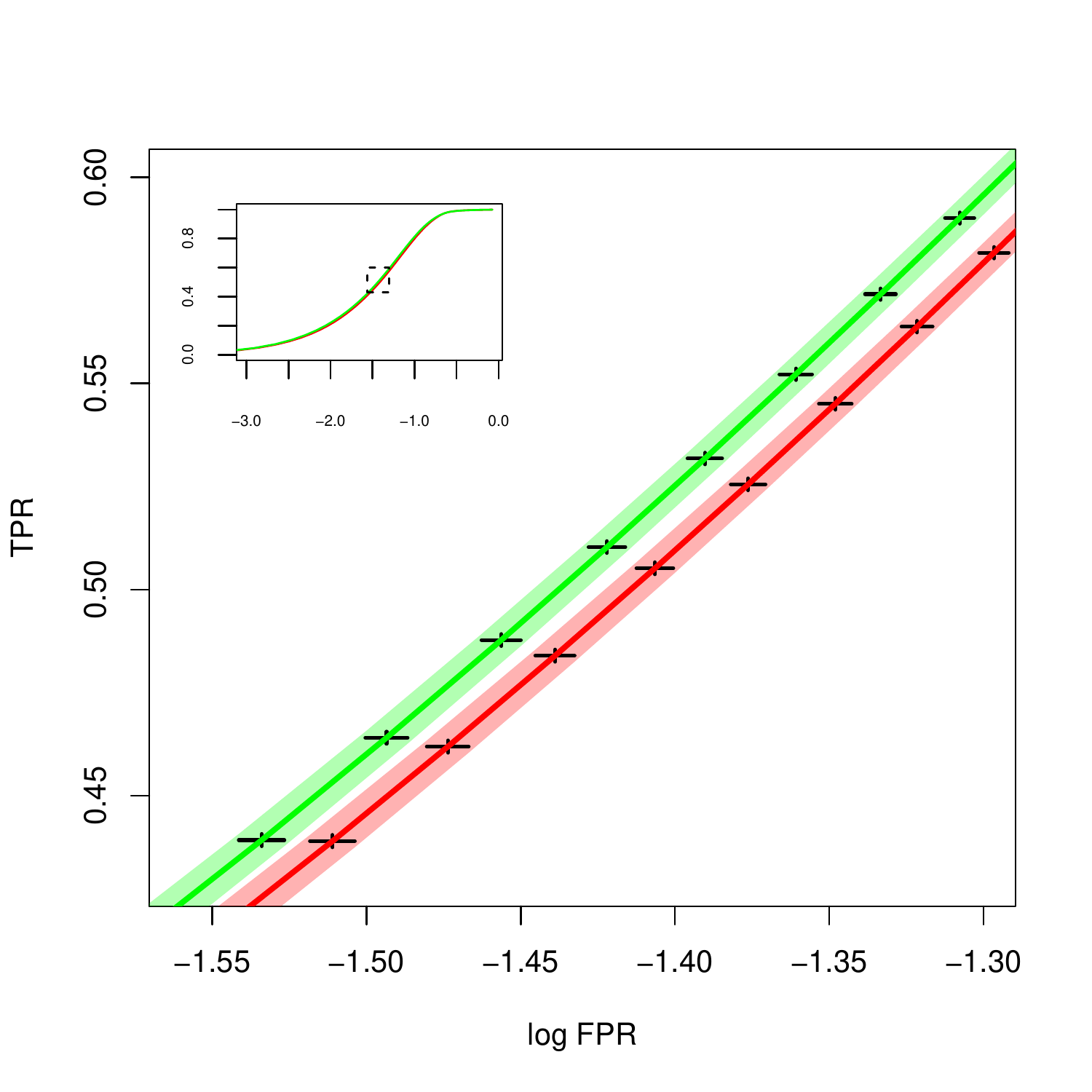}
  \caption{Confidence bands for random forests trained by
    \red{considering 64 features} and \green{considering 256 features}
    when splitting each node.  Also shown are two standard deviation
    confidence intervals (black) for log FPR and TPR at several points
    along the curve. The inset shows the full ROC curves. }
  \label{RF_more_features}
\end{center}
\end{figure}

Breiman \cite{Breiman} recommended trying $\sqrt{d}$ features to split
each node where $d$ is the number of features in the data set.  Some
widely-used implementations of random forests (such as {\tt sklearn})
do this by default.  Figure~\ref{RF_more_features} compares the
confidence bands for random forests that consider 64 features and 256
features to split each node.  Since the image data set has 4096
features, considering 64 features is standard.  However, considering
256 features clearly performs better.

\section{Conclusions and Ongoing Work}
\label{conclusions}

We have provided a simple, computationally efficient technique for
constructing ROC curves and associated confidence bands for ensemble
methods. This technique could easily be extended to deal with weak
learners producing non-binary output and to regression (as opposed to
classification problems) and other loss functions. The technique could
also be modified to deal with different aggregation techniques, by
considering non-uniform weights on collections of weak learners.  None
of these extensions is considered here, but the bones of the technique
are strong enough to support discussion and use.

We have demonstrated that there are at least two sources of
variation which influence the estimated operating characteristics 
of a given ensemble method: the particular weak learners included
in the ensemble and the particular training data set used to 
produce the estimates. It is important to think about both
sources of error when comparing techniques or assessing the
performance of a single technique.

Finally, we applied the technique to various implementations of the
random forest algorithm. This produced many insights, among them:
considering data resampling schemes other than the bootstrap can lead
to forests with improved predictive accuracy; and artificially
restricting the depth of individual trees or otherwise biasing their
output will lead to worse than default performance.  There are
infinitely more experiments that could be carried out, but doing so
would mean that this paper would never end.

\section*{Acknowledgments}

We would like to thank Larry Carter, Skip Garibaldi, Kyle Hofmann,
Doug Jungreis, Dan Mauldin, and Kartik Venkatram for helpful comments
and suggestions.

\bibliographystyle{plain}

\end{document}